\documentclass[submission,copyright,creativecommons]{eptcs}
\providecommand{\event}{FMAS 2024} 

\usepackage{amsmath,amsfonts}
\usepackage{algorithmic}
\usepackage{algorithm}
\usepackage{array}
\usepackage{textcomp}
\usepackage{stfloats}
\usepackage{url}
\usepackage{verbatim}
\usepackage{graphicx}
\usepackage{caption}
\usepackage{subcaption}
\usepackage{cite}
\usepackage{pgfplots}
\usepackage{xspace}
\usepackage{xcolor}
\usepackage{tikz}
\usepackage{listings}
\usepackage{xcolor}

\lstdefinelanguage{yaml}{
    keywords={true, false, null, yes, no, name, priority, error_action},
    keywordstyle=\color{green}\bfseries,
    basicstyle=\ttfamily,
    sensitive=false,
    comment=[l]{\#},
    commentstyle=\color{gray}\ttfamily,
    morestring=[b]",
    morestring=[b]',
    stringstyle=\color{red},
}

\definecolor{canva_yellow}{HTML}{FAC12C}
\definecolor{canva_green}{HTML}{00BF63}
\definecolor{canva_red}{HTML}{FF3131}
\definecolor{canva_blue}{HTML}{004AAD}

\usepackage{amsmath} 
\usetikzlibrary{arrows.meta, patterns}

\newcommand{\absname}{RV4Chatbot\xspace}
\newcommand{\rvrasa}{RV4Rasa\xspace}
\newcommand{\rvdialog}{RV4Dialogflow\xspace}

\allowdisplaybreaks

\usepackage{iftex}

\ifpdf
  \usepackage{underscore}         
  \usepackage[T1]{fontenc}        
\else
  \usepackage{breakurl}           
\fi

\title{\absname: Are Chatbots Allowed\\to Dream of Electric Sheep?}
\author{Andrea Gatti \quad \quad Viviana Mascardi
\institute{Department of Informatics, Bioengineering,\\ Robotics and Systems Engineering\\
University of Genoa \\ Genoa, Italy}
\email{forename.surname@unige.it}
\and
Angelo Ferrando
\institute{Department of Physics, Informatics \\and Mathematics\\
University of Modena and Reggio Emilia\\
Modena, Italy}
\email{angelo.ferrando@unimore.it}
}
\def\titlerunning{\absname: Are Chatbots Allowed to Dream of Electric Sheep?}
\def\authorrunning{A. Gatti, V. Mascardi \& A. Ferrando}
\begin{document}
\maketitle

\begin{abstract}
Chatbots have become integral to various application domains, including those with safety-critical considerations. As a result, there is a pressing need for methods that ensure chatbots consistently adhere to expected, safe behaviours. In this paper, we introduce \absname, a Runtime Verification framework designed to monitor deviations in chatbot behaviour. We formalise expected behaviours as interaction protocols between the user and the chatbot.
We present the \absname design and describe two implementations that instantiate it: \rvrasa, for monitoring chatbots created with the Rasa framework, and \rvdialog, for monitoring Dialogflow chatbots. Additionally, we detail experiments conducted in a factory automation scenario using both \rvrasa and \rvdialog. 
\end{abstract}

\section{Introduction}
\label{sec:introduction}

On November 30th, 2022, ChatGPT was unveiled \cite{chatgpt} deeply shaking the industry and academic worlds. The impression, at that time, was that ChatGPT and chatbots based on Large Language Models (LLM) would have irreversibly changed the way chatbots were designed and built,  wiping away any pre-existing technology. 

After almost two years, a more balanced view on the future is emerging, with the shared feeling that there is still room for chatbots that do not rely on generative AI techniques.  

There are many ways to classify chatbots based for example on the knowledge domain, the service provided, the goals, the response generation method \cite{ADAMOPOULOU2020100006}, the locus of control (chatbot- or user-driven) and duration of the interaction (short or long) \cite{DBLP:conf/insci/FolstadSB18}, or their affordances and disaffordances \cite{LIN202245,DBLP:journals/ce/JeonLC23}. For the purposes of this paper, the simplest and most suitable classification of text-based chatbots divides them into conversational AI and generative AI ones \cite{ibmchatbottypes}.

DialogFlow \cite{dialogflowsite,sabharwal2020introduction}, Rasa \cite{rasasite,DBLP:journals/corr/abs-1712-05181}, Wit.ai \cite{witaisite,mitrevski2018getting}, just to name a few, are text-based conversational AI chatbots, also referred to as intent-based chatbots. They can understand the users questions, no matter how they are phrased, thanks to Natural Language Understanding (NLU) capabilities that allow them to detect the user's intent and further contextual information. The NLU component exploits machine learning techniques for the intent classification and performs well even with a small amount of training sentences. The answer to be provided to the user is not autonomously generated by the chatbot, but is designed by the chatbot's developer. Conversational AI chatbots can remember conversations with users and incorporate contextual information into their interactions.

ChatGPT, Gemini \cite{geminisite}, Jasper Chat \cite{jaspersite} are examples of generative AI chatbots. They go far beyond conversational AI chatbots thanks to their capability of generating new content as their answer in form of high-quality text, images and sound based on LLMs they are trained on. This impressive power, however, does not come without   pitfalls. Besides religious bias \cite{DBLP:conf/aies/AbidF021}, gender bias and stereotypes \cite{DBLP:conf/ci2/KotekDS23}, and hallucinations  \cite{zhang2023sirens}, 
major privacy concerns are associated with LLMs. 

In March 2023, Italy's data regulator imposed a temporary ban on ChatGPT due to concerns related to data security. During its development, in November 2023, an open letter was signed by nine Italian scientific associations including the Italian Association for AI and the Italian Association for Computer Vision, Pattern Recognition and Machine Learning, and by around 500 scientists, asking the Italian government to guarantee that strict rules for the use of generative AI were included in the European AI Act \cite{AIAct}.

Scientific studies on LLM privacy leakage are so recent to be still unpublished at the time of writing, but many pre-prints by academic scholars show that the problem is real \cite{yang2023shadow,yong2023lowresource,chen2023janus}.   
Personal Identifiable Information (PII) protection can only be complied with by organisations able to have a private installation of a LLM within a private cloud or on premise \cite{conversationalUIs}. The resources needed to implement this solution make it not affordable for most companies and universities.

The global LLM market size (that includes the generative AI chatbot market plus a wide range of other applications) is projected to reach 259,886 Million USD revenue by 2029 \cite{gii}, while the conversational AI market is expected to reach 29,800 Million USD by 2028 \cite{cAIm}: the market forecasts and the privacy, ethical, and economical issues of LLM suggest that traditional conversational AI chatbots will still be needed and used by many players in the next few years. 

Although more controllable than their generative evolution, the behaviour of conversational AI chatbots can also be unsafe. In a factory automation scenario, where an intent-based chatbot provides a natural language interface between the user and a virtual representation of a factory, a conversation becomes unsafe if the user requests to position an object where another object has already been placed, or if the distance between objects is insufficient. Similarly, a conversation is unsafe if the chatbot provides the coordinates of an object that the user never inserted.

To cope with safety issues in conversational AI chatbots, we present an approach to verify at runtime the conversation between the user and the chatbot. Runtime Verification (RV)~\cite{DBLP:series/lncs/BartocciFFR18} is a formal verification technique used to analyse the runtime behaviour of software and hardware systems concerning specific formal properties. A RV monitor emits boolean verdicts that state whether the property is satisfied or not by the currently observed events. The default functioning is to state that something went wrong when it just went wrong, and trigger {\em recovery} actions. In some cases, the monitor may intervene before the wrong event is generated or the unsafe action is done, hence allowing for {\em prevention}. With respect to other formal verification techniques, such as Model Checking~\cite{clarke1997model} and Theorem Provers~\cite{DBLP:books/lib/Loveland78}, RV is more dynamic and lightweight and shares some similarities with software testing, being focused on checking how the system behaves while it is running. 

To perform RV of chatbots, we have designed a general and formalism-agnostic framework named \absname. We show \absname versatility by instantiating it for RV of chatbots created with Rasa, widely used to develop chatbots in local environments, and Dialogflow, used in cloud-based applications. 
We demonstrate how our engineering decisions render \absname a highly practical methodology and how it can seamlessly integrate with existing chatbot frameworks. It is essential to note that our utilisation of Rasa and DialogFlow serves only to illustrate potential applications of our approach. Our ultimate aim is to encompass any chatbot development framework. 


The paper is structured as follows. 
After overviewing the related work in Section~\ref{sec:related-work}, Section~\ref{sec:motivation-example} introduces one example that motivates the need of \absname. After that, Section~\ref{sec:foundation} describes the architecture and the data and control flow of \absname. 
Sections ~\ref{sec:rvrasa} and ~\ref{sec:rvdialog} describe, respectively, \rvrasa and \rvdialog, the two concrete instantiations of the \absname logical architecture. 
Section~\ref{sec:case-studies-and-exp} discusses the formalisation of some relevant safety properties in the motivating scenario, using a highly expressive RV language, and the experiments we carried out to verify those properties with \rvrasa and \rvdialog.
Section~\ref{sec:conclusions-future-work} concludes the paper and highlights the possible future directions.


\section{Related Work}
\label{sec:related-work}

RV of interaction protocols attracted the attention of researchers starting from the beginning of the millennium. The first interaction protocols to be verified at runtime involved web services \cite{li2006runtime}, cloud applications \cite{shao2010runtime},  cryptography \cite{bauer2010runtime}. RV of interactions among autonomous software agents followed soon\cite{alotaibi2010runtime,DBLP:conf/aciids/BakarS13}\footnote{Starting from 2012, a large share of the scientific production of the authors dealt with RV of agent interaction protocols, see \url{https://rmlatdibris.github.io/biblio.html}. We limit ourselves to cite the most relevant works among these.}.

Despite the large interest in RV of interactions and the pressing need to monitor what chatbots say and do, to the best of our knowledge no studies on RV of human-chatbot interactions exist, if we exclude the very recent works where we were involved. 
%
%

Apart from~\cite{DBLP:conf/vortex/0001GM23} which serves as the foundation for this paper but is tailored for a specific chatbot framework, the only other work in the literature that deals with the formal verification of chatbot systems is~\cite{engelmann2023rv4jaca}, introducing a framework known as RV4JaCa. In that work we integrated RV within the multiagent system (MAS) domain and demonstrated how to monitor agent interaction protocols within that context. The focus there was not on the chatbot itself, but on the software agents interacting with it. The main contribution was hence in the MAS domain, although applied in a scenario where messages for agents are generated by a chatbot.

Expanding the boundaries of our investigation, we can mention a recent proposal that approaches formal verification of chatbots from a static perspective, instead of at runtime as we do.
In~\cite{Ramos23} the authors introduce a strategy for verifying chatbot conversational flows during the design phase using the UPPAAL tool \cite{DBLP:conf/hybrid/BengtssonLLPY95}, a well-known model checker. The approach is tested by designing a hotel booking chatbot and receiving feedback from developers. The strategy is found to have an acceptable learning curve and potential for improving chatbot development.
%
%
In contrast to our approach, the work presented in~\cite{Ramos23} focuses on abstracting the chatbot using a model and subsequently verifying it through model checking. Due to the distinct inherent natures of these two verification approaches, we envision the possibility of integrating them to harness their respective strengths. Specifically, our technique could enhance the visibility of~\cite{Ramos23} by providing information that is only available at runtime. Conversely, the exhaustiveness of~\cite{Ramos23} could be leveraged by our approach to simplify the properties for monitoring, thanks to prior knowledge of the chatbot's behavioural model.



If we further expand our search and give up formality, hence resorting to software testing of chatbots, some works from J. Bozic's  research group can be mentioned. 
The paper~\cite{DBLP:conf/aitest/BozicTW19} introduces a planning-based testing approach for chatbots, focusing on functional testing, specifically in the context of tourism chatbots for hotel reservations. Planning is used to generate test scenarios, and a testing framework automates the execution of test cases. The results show success in testing chatbots, but some issues, such as intent recognition errors, need further attention. 
Metamorphic testing is illustrated in~\cite{DBLP:conf/pts/BozicW19}, where metamorphic relations are used instead of traditional test oracles due to the unpredictable nature of AI systems. 
On a similar line of research, the work~\cite{DBLP:journals/sqj/Bozic22} introduces an approach that leverages ontologies to generate test cases and addresses the absence of a test oracle by using a metamorphic testing approach. The method is demonstrated on a real tourism chatbot.

A methodology that automates the generation of coherence, sturdiness, and precision tests for chatbots and leverages the test results to enhance their precision is presented in~\cite{DBLP:conf/quatic/Bravo-SantosGL20}. The methodology is implemented in a tool called Charm, which uses Botium \cite{botium} for automated test execution. The paper also presents experiments conducted to improve third-party-built DialogFlow chatbots.

While these works share similarities with ours in that they focus on the actual, runtime execution of the chatbot rather than its abstraction, none is based on a rigorous and formal specification that guides the correctness checks.

To conclude, we emphasise that our goal is to assess the overall correctness of a conversation between a user and a chatbot as a coherent whole, rather than focusing on how individual message utterances are generated by the chatbot itself. This distinction is pivotal and sets our work apart from existing literature on the formal verification of Machine Learning models, as surveyed in~\cite{DBLP:conf/atva/SeshiaDDFGKSVY18}. In such literature, the emphasis is generally on verifying the Machine Learning model. In contrast, in \absname, our focus is not on whether the model correctly produces or classifies individual messages, but rather on the consistency of these messages within a conversation. In this sense, \absname delves deeper into the conversational semantics of the messages exchanged with or generated by the chatbot, rather than attempting to dissect the chatbot to understand its internal behaviour, which is often treated as a black-box.

\section{Motivating Example}
\label{sec:motivation-example}


Chatbots can be exploited for achieving three main goals: providing specific information stored in a fixed source (information chatbots); holding a natural conversation with the user (chat-based chatbots); and understanding the tasks that the user wants to perform, hence executing functions to perform them (task-based chatbots) \cite{ADAMOPOULOU2020100006}.

Usually, information chatbots provide an answer to one questions and go back to a state -- the only state they can be -- where they are ready to answer a new question. There is no need for them to keep memory of what the user already asked or said, and to carry out a coherent and fluent conversation. The correctness of the chatbot is related with the correctness of the search engine in its backend. Given that we aim at verifying the conversation flow rather than the quality of the retrieved information, RV of information chatbots following our approach is out of our scope.

Chat-based and task-based chatbots, on the other hand, engage into conversations that should evolve in different ways depending on what the user utters. For example, a chat-based chatbot may show different reactions to the very same request from the user, depending on how much the user insists upon it. In a task-based chatbot, the possibility for the user to ask for some task to be performed may depend on the fact that some prerequisite task had been asked, and hence performed, before.

Without loss of generality, the following motivating example focuses on a task-based chatbot. The application of \absname to chat-based chatbots is left for future work, as it would mainly require adapting the types of properties to be verified, rather than altering the verification methodology. In essence, \absname is not restricted to task-based chatbots; its architecture is sufficiently flexible to support the verification of any intent-based chatbot.

\paragraph{A Task-based Chatbot in the Factory Automation Domain.}
{This example, presented in the VORTEX 2023 workshop~\cite{DBLP:conf/vortex/0001GM23} and briefly summarised here, is set in the field of robotics and involves the development of a task-based chatbot assisting in the creation of a simulated factory work floor. The chatbot's role is to guide users through this process, taking into account both the users' requirements and the factory regulations\footnote{See ISO 10218-1:2011 standard on Robots and robotic devices - Safety requirements for industrial robots~\cite{ISO10218}.} concerning what can or cannot be added or removed from the factory work floor for safety reasons. The user interacts with the chatbot by requesting to add a robot to a specific position on the factory work floor, removing a robot, or relocating a previously added robot to a different position.
For example, properties verified at runtime might include ensuring that objects are not added to an already occupied position on the factory floor, confirming that each removal request corresponds to an object that actually exists in the current state, and enforcing spacing rules between objects as defined by safety regulations. The validity of these actions depends on the state of the simulated work floor, which evolves as the user-chatbot conversation progresses. RV4Chatbot checks these properties dynamically to detect and prevent any violations that could compromise the safety or coherence of the conversation.}

\section{\absname: The Foundation}
\label{sec:foundation}

\begin{figure*}[ht]
    \centering
    \includegraphics[width=1.0\textwidth]{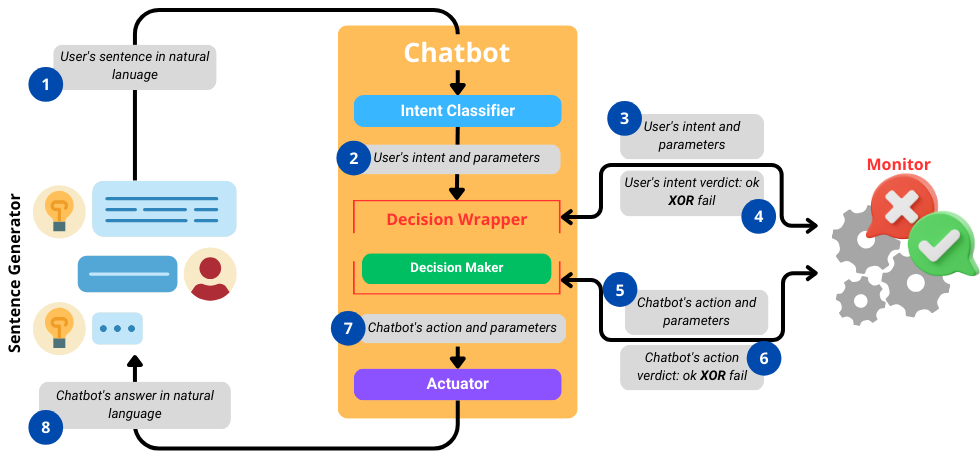}
    \caption{\absname architecture.}
    \label{fig:rv4chatbot-overview}
\end{figure*}

Figure~\ref{fig:rv4chatbot-overview} illustrates the operation of an intent-based chatbot. \absname specifically focuses on verifying intent-based chatbots, leaving the verification of non-intent-based chatbots for future work, as mentioned in both the introduction and conclusion sections.

A human user, or more generally, a sentence generator, produces a sentence in natural language (1). This sentence is categorised by a classifier based on its intent, and its parameters are extracted. The intent classifier is responsible for Natural Language Understanding (NLU). The recognised intent, along with its parameters (2), is then passed to a decision maker that determines the chatbot's response (7) to the input sentence. This decision-making process is typically hard-coded and integrated directly into the chatbot framework. Finally, the chatbot generates the response to be delivered to the user (8). The module responsible for this generation process is termed the `actuator'.

\absname introduces a decision wrapper, depicted in red with right angles inside the chatbot architecture, to manage actions numbered (3) to (6) on the right side of the figure. The decision wrapper extends the decision maker to allow the data characterising a chatbot's lifecycle—user's intents and chatbot's actions, both with optional parameters—to be sent to an external monitor where Runtime Verification (RV) occurs. Depending on the chatbot's framework and its modularity, implementing the decision wrapper may vary in complexity and intrusiveness. The decision wrapper instruments the `System Under Scrutiny' (the chatbot in this application), using standard RV terminology. In \absname, instrumentation is confined solely to this module.

The decision wrapper sends the recognised intents and parameters (3) to the monitor. Regardless of its implementation and the language used for modeling properties to verify, the monitor observes one event at a time and emits a boolean verdict indicating whether the event complies with the property (4). If the verdict is true (or inconclusive\footnote{We remind the reader that in RV, it is common to have at least a third outcome indicating that the monitor does not yet have sufficient information to determine whether the property under analysis is satisfied or violated.}), control returns to the decision maker, which decides the subsequent action. Before executing the action, the decision wrapper sends it to the monitor (5), which again verifies its compliance with the property and emits a verdict (6).

A true (or inconclusive) verdict from the monitor does not alter the chatbot's standard execution flow. A false verdict—whether originating from an unexpected user intent or a disallowed action by the chatbot—returns the chatbot to a listening state, displaying a message explaining the failure to the user. In both cases, no unsafe actions are performed.

Numerous intent-based chatbot frameworks are documented in the literature. Although their implementations may vary significantly, their main components and functionalities are accurately represented in Figure~\ref{fig:rv4chatbot-overview}. Similarly, many RV monitors exist. Regardless of the monitor used, it must at least be able to observe events from the System Under Scrutiny and output a verdict that is either true, false, or inconclusive. This is the only assumption we make regarding the RV monitor's function, and it is satisfied by the definition of a monitor. Thus, the \absname logical architecture is parametric in both the chatbot framework and the monitor.

To automate the experiments presented in Section \ref{sec:experiments}, we developed a piece of software capable of reading natural language sentences from a file and sending them to the chatbot using the APIs provided by the chatbot frameworks considered in this paper. Although our primary interest lies in RV, we soon realised that the files of simulated user sentences could be seen as test cases, and that the software component named `sentence generator' in Figure~\ref{fig:rv4chatbot-overview} (left side) could be used to run batches of tests. We re-engineered this component and elevated it to the status of one of the \absname components. This approach allows testing the chatbot during its development by exploiting the monitor as an offline test engine. The advantage of this method is that once the chatbot has been tested offline and then deployed, the monitor can continue to function at runtime, in line with its primary objective. No code changes are required in the monitor or the instrumented chatbot when switching from offline testing to RV; only the source of sentences changes, becoming a human user in the latter case.

Now that we have completed the introduction of \absname, we can focus on its two instantiations for the RV of Rasa and Dialogflow chatbots.

\section{\rvrasa}
\label{sec:rvrasa}


\subsection{Rasa}

``\textit{Rasa Open Source is an open source conversational AI platform allowing developers to understand and hold conversations, and connect to messaging channels and third party systems through a set of APIs.}"\footnote{\url{https://rasa.com/docs/rasa/}}

Rasa \cite{DBLP:journals/corr/abs-1712-05181} is composed by two different tools: Rasa NLU and Rasa Core.
When a message is received from the user, Rasa NLU extracts the intent and the entities (namely, structured pieces of information inside a user message) from it.
The structured information is then passed to the Tracker object. This object is used to store the dialogue state.
The tracker object is then passed to the policies.
Each policy has a ranking and can return a list containing one score for every possible action to perform next.
Rasa Core will perform the action with the best score provided by the highest ranked policy.
The action server executes the action, the tracker object is updated and then passed again to the policies.
When no more actions to be performed are available, the policies return the `listen' action, waiting for a user input.

Rasa NLU and policies use three files for training:
\begin{itemize}
    \item \texttt{nlu.yml}, containing all the example sentences uses to train the intent classification and the entities extraction;
    \item \texttt{stories.yml}, containing all the paths that a conversation can follow;
    \item \texttt{rules.yml}, containing stricter conversation patterns and actions that must take place if triggered.
\end{itemize}
Rasa Core employs two main files for the flow control and configuration:
\begin{itemize}
    \item \texttt{domain.yml}, containing all the information on what Rasa NLU can extract (intents and entities), definition of slots (stored values), available actions and responses;
    \item \texttt{config.yml}, containing the pipeline for the Rasa NLU training and the policies definition.
\end{itemize}
Rasa actions can be defined either as strings to answer or as complete Python classes to be executed when called.

\begin{figure*}[ht]
\centering
    \includegraphics[width=1.0\textwidth]{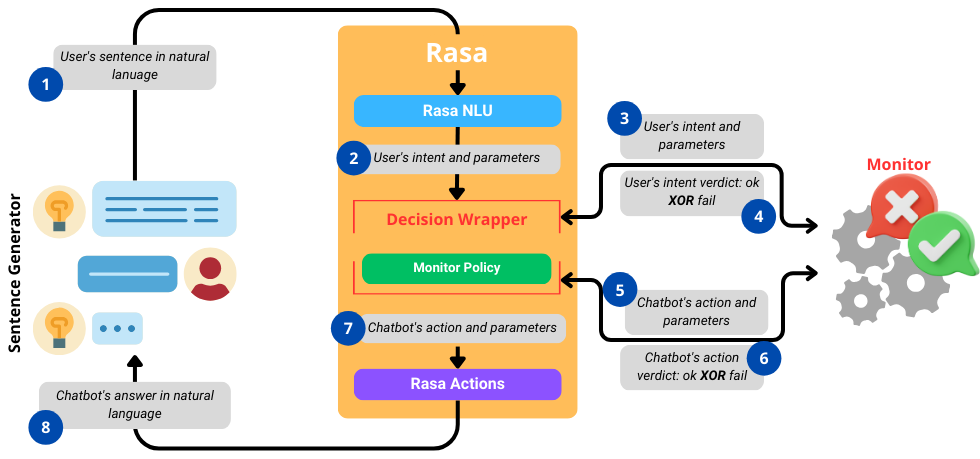}
    \caption{The \rvrasa instantiation of \absname.}
    \label{fig:rv4chatbot-overview-rasa}
\end{figure*}

\subsection{The \rvrasa instantiation of \absname}

The Rasa architecture aligns closely with that of \absname, as shown in Figure~\ref{fig:rv4chatbot-overview-rasa}. Each element of the \absname architecture maps directly to a specific component in the \rvrasa instantiation, without requiring any modifications to Rasa’s original design.

The Sentence Generator can be either the human user using the shell provided by Rasa or a script that sends messages as \textsc{post} requests to the Rasa server provided by Rasa.
The Intent Classifier is the Rasa NLU that extracts intent and entities.
The Decision Wrapper in Rasa is managed by the policies.
In particular, for this module it is necessary to add a policy (\texttt{monitorPolicy}) that sends an event to the monitor for each action executed.
The Actuator overlaps with the Rasa Actions that can perform any piece of provided Python code.

Notice that the policies predict the next action based on the previous ones so the \texttt{monitorPolicy} can only stop the chatbot immediately after the wrong action has been executed. This paves the way to {\em recovery} from wrong actions, but not to {\em prevention}. However, by exploiting Rasa policies the developer only needs to add the \texttt{monitorPolicy} to them, without any other change to the chatbot; RV will be performed automatically thanks to the \texttt{monitorPolicy}. The simplicity and the minimal invasiveness of `injecting' RV capabilities into Rasa this way, motivates our decision to give up prevention, and accept that the monitor realizes that something went wrong, after this already happened. Actually, ex-post notification is a standard operating way in RV.  

\subsection{Challenges in the \rvrasa design and development}

The main effort required by the \rvrasa development was understanding how policies work, and implementing the \texttt{monitorPolicy}.
In fact, whereas Rasa's documentation is extensive and well-assorted for a basic usage, it is almost completely absent when policies come into play.
The policy is added in the config file as follows:
% ...
\begin{lstlisting}
policies:
...
  - name: policies.monitorPolicy.MonitorPolicy
    priority: 6
    error_action: "utter_error_message"
\end{lstlisting}

In its implementation the main class, \texttt{monitorPolicy}, inherits Rasa's Policy class; in particular, it inherits and redefines:
\begin{itemize}
    \item \texttt{\_\_init\_\_}, the initialisation method, here the error action provided by the user is saved or set to a default value if needed;
    \item \texttt{predict\_action\_probabilities}, called every time the policy runs and returning the list of probabilities. This method may also return no value at all, and this is exactly the way we use it, to keep the conversation flowing as if no RV were performed, if no errors occur.
\end{itemize}
Note that, the \texttt{priority} assigned to the policy is user-defined and ensures that Rasa gives precedence to the \texttt{monitorPolicy} over other custom policies. This higher priority is crucial since the \texttt{monitorPolicy} addresses safety aspects, which must take precedence in the chatbot’s decision-making process.

\subsection{Source Code}

To instantiate \rvrasa there are only two additions to be made in the chatbot:
\begin{enumerate}
    \item \texttt{monitor\_policy.py}: ~150 lines of code;
    \item \texttt{config.yml}: 3 further lines should be added to the configuration file, to turn Rasa into \rvrasa.
\end{enumerate}

The code of \rvrasa is available at \url{https://github.com/driacats/RV4Chat/tree/main/Rasa}.

\section{\rvdialog}
\label{sec:rvdialog}


\subsection{Dialogflow}
Dialogflow \cite{dialogflow} is a lifelike conversational AI platform developed by Google that enables users to create virtual agents equipped with intents, entities (similar to those in Rasa), and fulfillment. Fulfillment refers to the capability of these agents to interact with external systems or APIs to retrieve dynamic responses, process data, or execute specific actions based on the user's input, going beyond pre-defined static responses.

Dialogflow performs NLU using \textit{intents}, defined via a name and a set of training example sentences.
Dialogflow trains a model able to identify, for each user message sent on the chat, the \textit{nearest} intent and the confidence score.
Training sentences may also contain \textit{entities}, namely pieces of information that may be significant for the conversation and that should be extracted from the text.
For each intent, a bunch of possible answers may be displayed. However, some messages cannot be answered from inside Dialogflow, as they require to process data or execute operations. In this case, users can use the \textit{fulfillment} for sending a message to an external server that will execute the correct actions, and provide an answer back.

\begin{figure*}[ht]
\centering
    \includegraphics[width=1.0\textwidth]{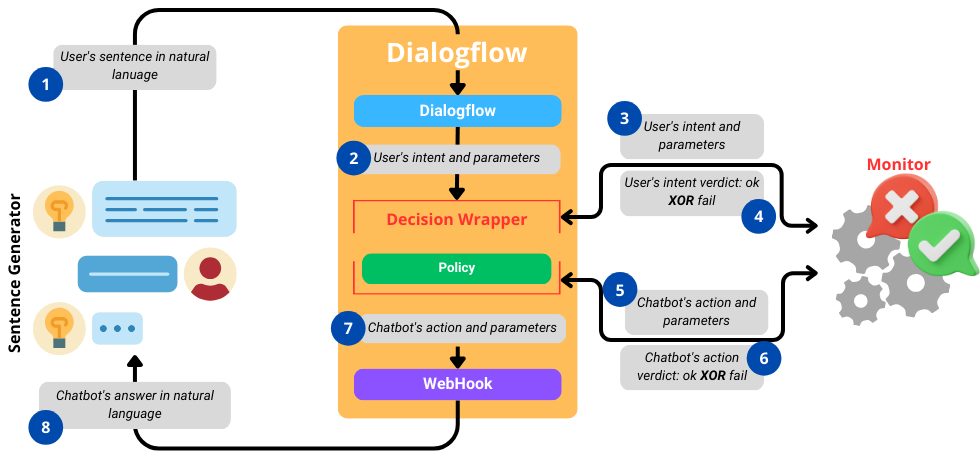}
    \caption{The \rvdialog instantiation of \absname.}
    \label{fig:rv4chatbot-overview-dialog}
\end{figure*}

\subsection{The \rvdialog instantiation of \absname}
\rvdialog is structured as shown in Figure \ref{fig:rv4chatbot-overview-dialog}.
To instantiate \absname in Dialogflow, we had to add a brand new component to the Dialogflow architecture. This made the design and implementation of \rvdialog much more complex than the \rvrasa one.

This additional component can be generated directly from an exported Dialogflow agent using an instrumentation script that we developed. We call this brand new component \textit{policy}, for analogy with \rvrasa.

The instrumentation script has two outputs: a .zip file containing the new Dialogflow agent, and the policy python script.
In the new Dialogflow agent, every message is forwarded to the policy that controls the flow. The policy maintains the original agent's flow, forwarding only the necessary messages back to DialogFlow through a webhook\footnote{Which is the mechanism used in DialogFlow to communicate to DialogFlow from an external service, that in \rvdialog is the policy.}. Additionally, for each message and action performed, the policy sends a message to the monitor.

\subsection{Challenges in the \rvdialog design and development}

\begin{figure}[ht]
    \centering
    \includegraphics[width=0.8\textwidth]{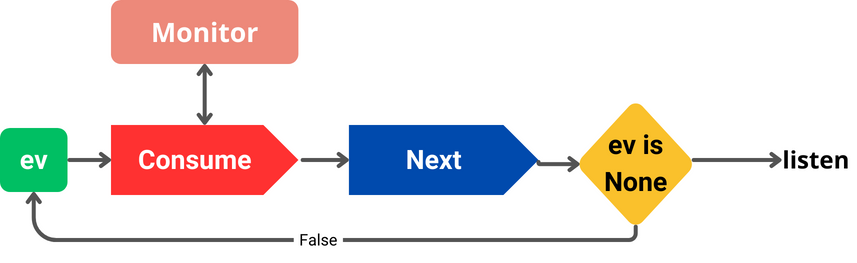}
    \caption{\rvdialog policy flow.}
    \label{fig:dialog-policy}
\end{figure}

The \rvdialog policy works with \textit{events} as shown in Figure \ref{fig:dialog-policy} (reported as ev).
There are two main types of events: user and bot events.

The policy can receive messages both from the user or the bot.
When it receives a message it creates an event.
An event in this domain can be of five main types: (1) a user message with all its features; (2) a bot message with all its features; (3) a plain answer to be sent as answer; (4) an action to be performed by the webhook; (5) the error action.

The event is then consumed: it is sent to the monitor and then if it is a plain answer it is sent on the chat, if it is an action it is performed.
Obviously, if the monitor claims an error the action is set immediately to the error one.

When the event is consumed the policy computes the next one.
For examples, if the event is a user message the next event is the answer.
The policy consumes and computes next event until the next event is None, in this case it listens for new inputs.

\subsection{Source Code}

To instantiate \rvdialog the needed files are:
\begin{enumerate}
    \item \texttt{instrumenter.py}: ~190 lines of code;
    \item \texttt{policy.py}: (generated by the instrumenter) from ~150 lines of code;
\end{enumerate}

The code of \rvdialog is available at \url{http://github.com/driacats/RV4Chat/tree/main/Dialogflow}.

\section{Experiments}
\label{sec:case-studies-and-exp}

The formalism we use to model properties to be verified at runtime is named Runtime Monitoring Language (RML) and has been selected for  its high expressive power that goes beyond Linear Temporal Logic (LTL) \cite{DBLP:conf/birthday/AnconaFM16} and the familiarity of the authors. 
As explained in Section \ref{sec:foundation}, the \absname framework is meant to be instantiated with any conversational chatbot framework and any RV language and tool. RML is one among the many existing RV languages and its  adoption is only functional to run experiments with \rvrasa and \rvdialog. 

In this section, we briefly introduce RML and illustrate the RML properties that capture safety requirements in the factory automation case study, providing the RML encoding of one of the properties verified in that scenario (the complete encoding can be found in~\cite{DBLP:conf/vortex/0001GM23}). All these properties are correctly verified by the monitor, so we do not allocate space to the qualitative experiments we conducted, as they can be summarised by stating that ``the monitor always works as expected''. Instead, we present performance experiments, demonstrating that the addition of the monitor to the chatbot introduces negligible overhead.

\subsection{Runtime Monitoring Language}

The Runtime Monitoring Language (RML~\cite{RMLsite, DBLP:journals/scp/AnconaFFM21}) is a Domain-Specific Language (DSL) for specifying highly expressive properties in RV such as non context-free ones. We chose to use RML in this work because of its support of parametric specifications and its native use for defining interaction protocols.

Since RML is just a means for our purposes, we only provide a condensed view of its syntax and denotational semantics in terms of the represented traces of events. A detailed explanation of some of its operators is provided in Section \ref{subsect:factoryautomationproperties} where RML specifications are provided. The complete presentation can be found in~\cite{DBLP:journals/scp/AnconaFFM21}.

In RML, a property is expressed as a tuple $\langle t,\mathit{ETs} \rangle$, with $t$ a term and $\mathit{ETs}=\{\;ET_1,\ldots,\allowbreak ET_n\;\}$ a set of event types. An event type $ET$ is represented as a set of pairs $\{\;k_1:v_1,\ldots,k_n:v_n\;\}$, where each pair identifies a specific piece of information ($k_i$) and its value ($v_i$). An event $Ev$ is denoted as a set of pairs $\{\;k_1':v_1',\ldots,k_m':v_m'\;\}$. Given an event type $ET$, an event $Ev$ matches $ET$ if $ET \subseteq Ev$, which means $\forall (k_i:v_i) \in ET \cdot \exists (k_j:v_j) \in Ev \cdot k_i = k_j \land v_i = v_j$. In other words, an event type $ET$ specifies the requirements that an event $Ev$ has to satisfy to be considered valid.

An RML term $t$, with $t_1$, $t_2$ and $t'$ as other RML terms, can be:
\begin{itemize}
    \item $ET$, denoting a set of singleton traces containing the events $Ev$ s.t. $ET \subseteq Ev$;
    \item $t_1 \;\; t_2$, denoting the sequential composition of two sets of traces;
    \item $t_1 \; | \; t_2$, denoting the unordered composition of two sets of traces (also called shuffle or interleaving);
    \item $t_1 \land t_2$, denoting the intersection of two sets of traces;
    \item $t_1 \lor t_2$, denoting the union of two sets of traces;
    \item $\{\; let \; x;\; t' \;\}$, denoting the set of traces $t'$ where the variable $x$ can be used (i.e., the variable $x$ can appear in event types in $t'$, and can be unified with values);
    \item $t' *$, denoting the set of chains of concatenations of traces in $t'$.
\end{itemize}

Event types can contain variables (we use the terms \emph{argument} and \emph{variable} interchangeably). 
%
In RML, recursion is modelled by syntactic equations involving RML terms, such as $t = ET_1  \; t \lor ET_2$, modeling a finite (possibly empty) sequence of events matching the event type $ET_1$ ended by one event matching  $ET_2$, or the infinite trace including only events matching $ET_1$.

\subsection{Factory Automation Domain properties}
\label{subsect:factoryautomationproperties}


The three properties to be verified in the factory automation domain have been presented in the VORTEX 2023 paper \cite{DBLP:conf/vortex/0001GM23}. We report one of them to better clarify the use of RML and the kind of protocols we are interested in verifying at runtime. 

The first property aims at ensuring that the user does not add an object in an already taken position. The corresponding RML specification is reported in the following (as in~\cite{DBLP:conf/vortex/0001GM23}).
{
\begin{align*}
AddObject &= \{\; let \;\; x,\; y; \\
& (msg\_user\_to\_bot \;\land\; add\_object(x, y)) \\
& (msg\_bot\_to\_user \;\land\; object\_added) \\
& (not\_add\_object(x, y)\!* \;\land\; AddObject)\; \;\} \\
ETs &= \{\; \;msg\_user\_to\_bot, \\
& msg\_bot\_to\_user, add\_object(x, y), \\
& object\_added \;\} \\
msg\_user\_to\_bot &= \{\; sender:\;``user", \;receiver:\:``bot" \;\} \\
msg\_bot\_to\_user &= \{\; sender:\;``bot", \;receiver:\:``user" \;\} \\
add\_object(x, y) &= \{\; intent:\; \{\; name:\; ``add\_object" \;\}, \\
& slots:\; \{\; horizontal:\; x, vertical:\; y \;\} \;\} \\
object\_added &= \{\; last\_action:\; ``utter\_add\_object" \;\}
\end{align*}
}

As an example, the user's request \emph{``Add a robot in position (3, 5)''} is safe only if the position \emph{(3, 5)} is empty. But, the position is empty if the user did not already ask to put objects there. Hence, the history of the previous interactions must be taken into account, to verify the feasibility of a new object addition. The property is parametric w.r.t. coordinates and is defined recursively; it involves the definition of four event types and exploits the $let$, $\land$, and $*$ RML operators. 

\begin{figure}[ht]
    \centering
	\begin{subfigure}{.75\textwidth}
	\centering
		\includegraphics[width=\textwidth]{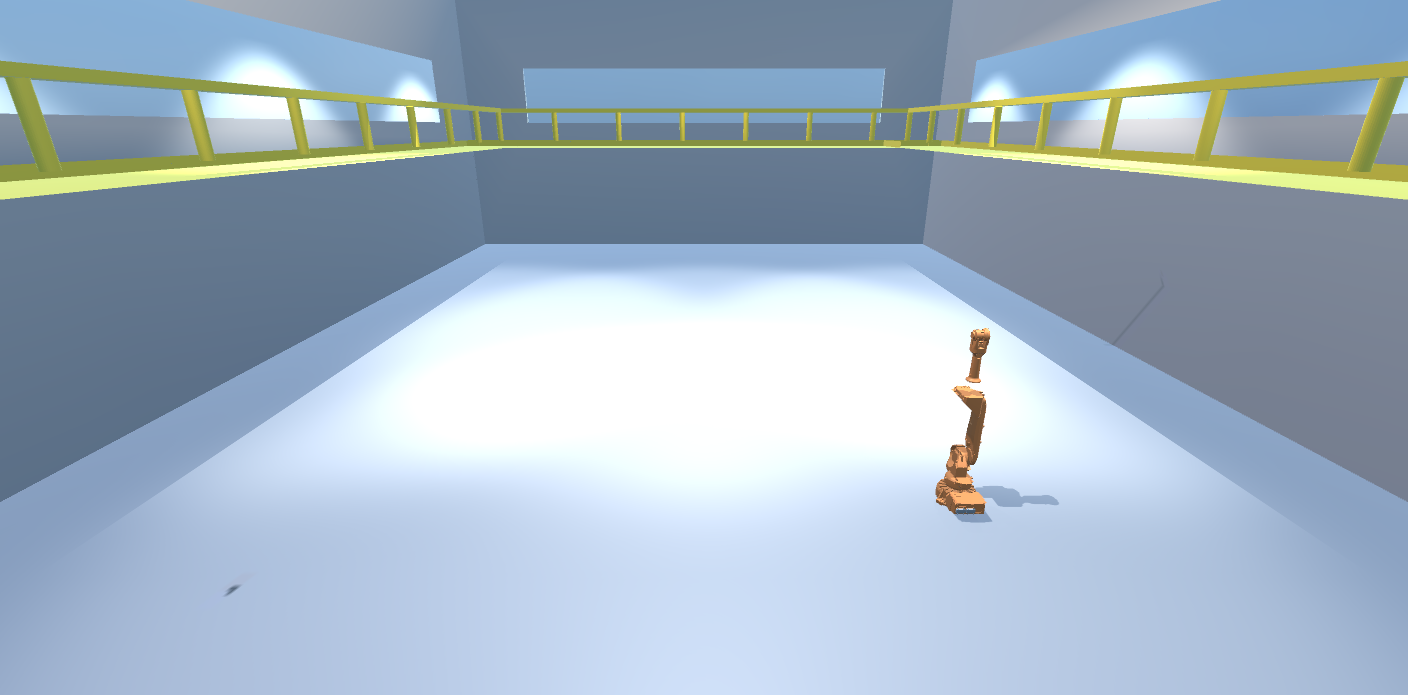}
	\end{subfigure}
	\begin{subfigure}{.75\textwidth}
	\centering
		\includegraphics[width=\textwidth]{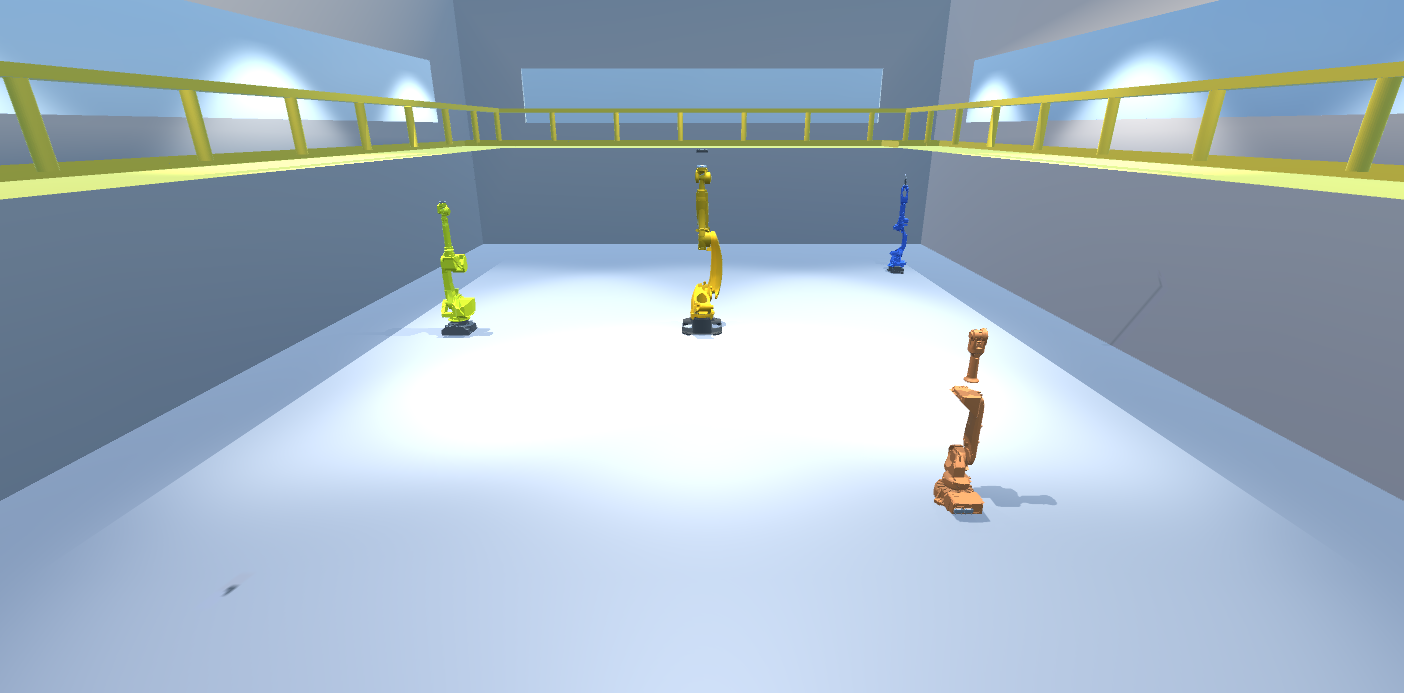}
	\end{subfigure}
    \caption{Initial scenario of the simulated factory floor (above) and the result after further iterations of adding new objects in the scene (below) taken from~\cite{DBLP:journals/robotics/GattiM23}.}
    \label{add-final}
\end{figure}

Figure~\ref{add-final} presents screenshots of the simulated environment in which the Rasa (and Dialogflow) chatbots operate. These screenshots specifically demonstrate how, by interacting with the chatbot, the user can add new objects to the simulated factory floor. This interaction occurs under the scrutiny of the RML monitor, which checks, among other things, the previously mentioned property.

The second property, whose RML encoding is more complex than the previous one, deals with the addition of an object in a position which is relative to another object in the simulation. The user's request may be \emph{``Add a robot on the left of Robot3''}. In order to be a safe request, \emph{Robot3} should have been previously positioned somewhere, hence previous messages involving added and removed objects, their name, and their position in the simulation must be taken into account. The property is parametric w.r.t. coordinates and objects names, and defined recursively; it involves eight event types and exploits the $|$ and $\lor$ RML operators, besides $let$, $\land$, and $*$. The $|$ operator is used, for example,  to cope with the interleaving of future additions relative to the currently added object and additions of objects that are not relative to the newly added one. Disjunction is used to discriminate between the situation where one added object is then removed, and hence no further references to it are allowed, and the situation where no removal takes place, and references are safe.  

The third property exploits the RML feature of constraining values in event types. It checks that any observed message has the value associated with its \textit{confidence} field -- as returned by the NLU component of the chatbot -- greater than 60\%. 

\subsection{Performance Evaluation}
\label{sec:experiments}

All the experiments can be tested using the code provided here \url{https://github.com/driacats/RV4Chat/tree/main/Examples}.
In particular there are three important scripts for each experiment:
\begin{itemize}
    \item \texttt{start\_service.py}: this script allows the user to select the platform (Rasa, Dialogflow), the monitor (no monitor, dummy monitor, real monitor), and the scenario (factory automation), and starts the service;
    \item \texttt{run\_test.py}: this script launches the test conversation. The input messages are stored in the \texttt{test\_input.txt} file, and the conversation is iterated a certain number of times for each combination of platform and monitor. For each iteration, a file is created to store the response times for each message sent;
    \item \texttt{chat.py}: this script provides a chat interface to test the chatbot. It takes as argument the platform and manages the connection automatically. To test the program it is sufficient to start the service and then launch the chat with the same platform.
\end{itemize}
The tests have been performed using \rvrasa and \rvdialog with three different monitoring levels:
\begin{enumerate}
    \item without a monitor;
    \item with a dummy monitor that replies always \texttt{True};
    \item with a real monitor that checks the properties discussed in the previous sections.
\end{enumerate}

For this experiment the chatbot can identify three intents and five entities.
The three intents are (1) add an object (2) add a object with a reference to another object (3) remove an object, while the entities are (1) object to add or remove (2) vertical position (3) horizontal position (4) relative position (5) reference object.

The Dialogflow WebHook in this case is more complex and manages the addition and the removal of objects inside a real virtual environment.
The implementation of this experiment in Rasa, with a real Virtual Environment in the backend and a Multi-Agent System in the middle, has been presented in \cite{DBLP:conf/vortex/0001GM23}. For the tests presented here, the API calls that in \cite{DBLP:conf/vortex/0001GM23} accessed the virtual environment are instead sent to a dummy script that provides a terminal based representation of a virtual space and simulates the execution. No virtual environment implementation is involved in this experiment, which is aimed at testing the performance of the RV mechanism. 
\begin{figure}[ht]
\centering
\includegraphics[width=0.8\linewidth]{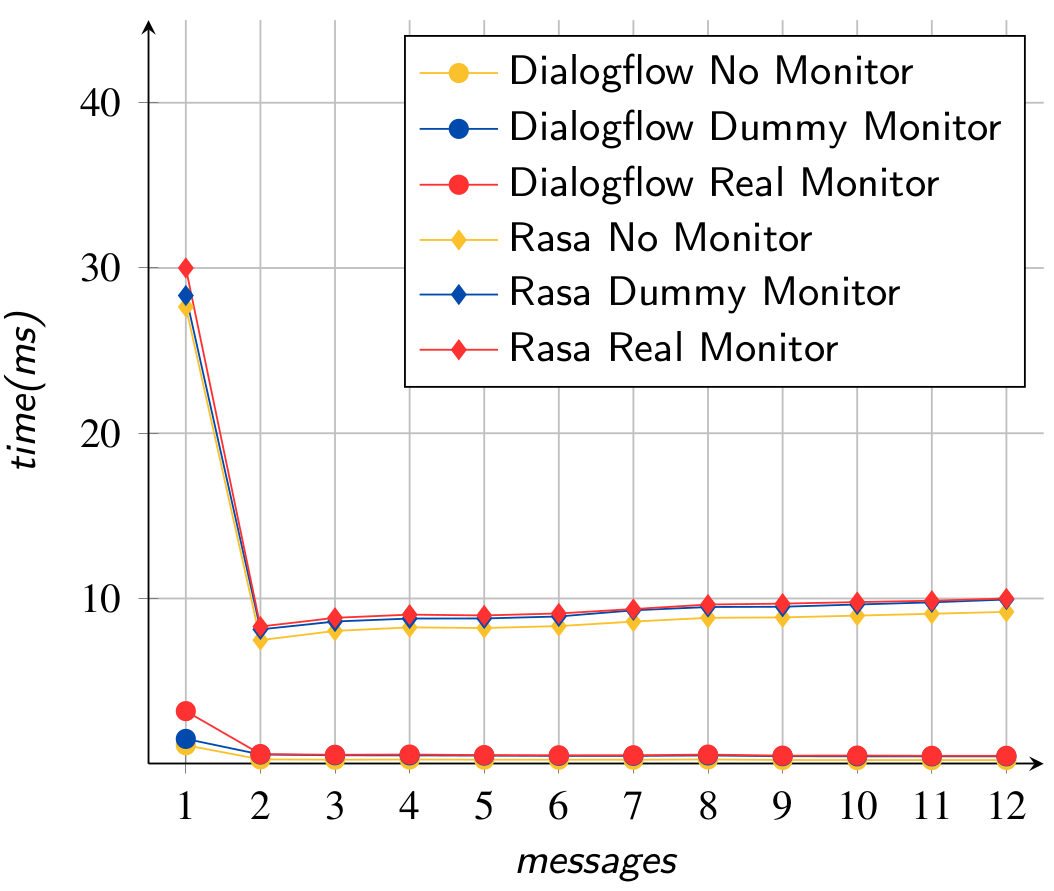}
\caption{Factory Automation Domain times on a test conversation of 12 messages. Messages for the test are: (1) \textit{Add a table} (2) \textit{Add a box right of table1} (3) \textit{Add a robot in front on the left} (4) \textit{Add a robot in front on the right} (5) \textit{Remove box0} (6) \textit{Remove robot1} (7) \textit{Add a table behind on the left} (8) \textit{Add a robot behind on the right} (9) \textit{Remove table1} (10) \textit{Remove table2} (11) \textit{Remove robot2} (12) \textit{Remove robot3}.}
\label{fig:factory-times}
\end{figure}

As shown in Figure \ref{fig:factory-times}, the monitor does not affect the execution time of the chatbot. The first message exhibits a significant time delay compared to the subsequent messages when using Rasa. This behaviour is due to Rasa itself: the Rasa Tracker object and all necessary instances for the conversation are initialised with the first message rather than at the server launch, resulting in a significantly higher time required to process the first message compared to the others.

\section{Conclusions and Future Work}
\label{sec:conclusions-future-work}


This paper introduces \absname, a framework for verifying the behaviour of conversational AI chatbots. \absname achieves this in a versatile manner, imposing minimal constraints on both the chatbot creation framework and the monitors deployed at runtime for formal verification. To demonstrate its efficacy, this paper presents two implementations of \absname: \rvrasa and \rvdialog. The engineering and experimental outcomes of these implementations are detailed, particularly when applied to safety-critical case studies in domains such as factory automation.

The experimental findings underscore \absname's generality, efficiency, and lightweight nature in terms of the overhead introduced by its monitoring components.

Looking ahead, our plans involve further exploration and experimentation with \absname, including its application to more complex case studies that can better challenge the framework's robustness and performance. This will allow us to assess how the performance overhead of \absname is impacted when applied to larger, real-world conversational systems with increased message volumes, more complex dialogue flows, and higher interaction frequencies. Additionally, while our current focus is on conversational AI chatbots, we plan to evaluate the scalability of \absname to understand how it performs as the number of monitored properties, intents, and concurrent conversations grows. Preliminary intuition suggests that the framework’s modularity may support scaling to moderately large applications, but this hypothesis needs to be tested empirically.

Furthermore, the insights and experiences gained from this work may facilitate future developments for handling generative chatbots. In such scenarios, where intents may be unavailable and decision-making is based on machine learning techniques, we aim to refine and expand \absname to integrate more dynamic monitoring approaches that can accommodate the unpredictability and complexity of generative AI.

\bibliographystyle{eptcs}
\bibliography{main}
\end{document}